\pdfoutput=1

\documentclass[11pt]{article}

\usepackage{emnlp2021}

\usepackage{times}
\usepackage{latexsym}

\usepackage[T1]{fontenc}

\usepackage[utf8]{inputenc}

\usepackage{microtype}

\usepackage{graphicx}
\usepackage{bm}
\usepackage{amssymb}
\usepackage{amsmath}
\usepackage[T1]{fontenc}
\usepackage{booktabs}
\usepackage{xcolor}
\usepackage{multirow}
\usepackage{booktabs}
\usepackage{verbatim}
\usepackage{mdwlist}
\usepackage[ruled,norelsize,linesnumbered]{algorithm2e}
\usepackage{amsthm}
\usepackage[capitalize]{cleveref}
\usepackage{paralist}
\usepackage{enumitem}
\usepackage{subcaption}
\usepackage{array}
\usepackage{amsfonts}
\usepackage{pifont}

\theoremstyle{definition}

\newcommand{\cond}{\,|\,}

\makeatletter
\newcommand{\removelatexerror}{\let\@latex@error\@gobble}
\makeatother

\usepackage{paralist}
\setitemize{itemsep=1pt,topsep=1pt,parsep=1pt,partopsep=1pt}

\usepackage{xspace}
\newcommand{\nmt}{\textsc{nmt}\xspace}
\newcommand{\mt}{\textsc{mt}\xspace}

\newcommand{\editor}{\textsc{editor}\xspace}
\newcommand{\cmlm}{\textsc{cmlm}\xspace}
\newcommand{\ar}{\textsc{ar}\xspace}
\newcommand{\nar}{\textsc{nar}\xspace}
\newcommand{\narc}{\textsc{nar+c}\xspace}

\newcommand{\tla}{\textsc{tla}\xspace}
\newcommand{\cd}{\textsc{cd}\xspace}
\newcommand{\bleu}{\textsc{bleu}\xspace}

\newcommand{\wmt}{\textsc{wmt}\xspace}

\title{Rule-based Morphological Inflection \\ Improves Neural Terminology Translation}

\author{Weijia Xu \\
	University of Maryland \\
	{\tt \href{mailto:weijia@cs.umd.edu}{weijia@cs.umd.edu}} \\\And
	Marine Carpuat \\
	University of Maryland \\
	{\tt \href{mailto:marine@cs.umd.edu}{marine@cs.umd.edu}} \\}

\begin{document}
\maketitle
\begin{abstract}
Current approaches to incorporating terminology constraints in machine translation (\mt) typically assume that the constraint terms are provided in their correct morphological forms. This limits their application to real-world scenarios where constraint terms are provided as lemmas. In this paper, we introduce a modular framework for incorporating lemma constraints in neural \mt (\nmt) in which linguistic knowledge and diverse types of \nmt models can be flexibly applied. It is based on a novel cross-lingual inflection module that inflects the target lemma constraints based on the source context. We explore linguistically motivated rule-based and data-driven neural-based inflection modules and design English-German health and English-Lithuanian news test suites to evaluate them in domain adaptation and low-resource \mt settings. Results show that our rule-based inflection module helps \nmt models incorporate lemma constraints more accurately than a neural module and outperforms the existing end-to-end approach with lower training costs.\footnote{Code and test suites are released at \url{https://github.com/Izecson/terminology-translation}}
\end{abstract}

\section{Introduction}
\looseness=-1
Incorporating terminology constraints in machine translation~(\mt) has proven useful to adapt translation lexical choice to new domains~\citep{HokampL2017} and to improve its consistency in a document~\citep{TureOR2012}.
In neural \mt~(\nmt), most prior work focuses on incorporating terms in the output exactly as given, using soft~\citep{SongZYLWZ2019,DinuMFA2019,XuC2021} or hard constraints~\citep{HokampL2017,PostV2018}. These approaches are problematic when translating into morphologically rich languages where terminology should be adequately inflected in the output, while it is more natural and flexible to provide constraints as lemmas as in a dictionary.

To the best of our knowledge, only one paper has directly addressed this problem for neural \mt: \citep{BergmanisP2021} design an \nmt model trained to copy-and-inflect the terminology constraints using target lemma annotations~(\tla) --- \tla are synthetic training samples where the source sentence is tagged with automatically generated lemma constraints. While this approach improves translation quality, the end-to-end training set-up prevents fast adaptation to lemmas and inflected forms that are rare or unseen at training time. Its impact is also limited to a specific neural architecture, and it is unclear whether its benefits port to more generic sequence-to-sequence models.

\looseness=-1
In this paper, we introduce a modular framework for inflecting terminology constraints in \nmt. It relies on a cross-lingual inflection module that predicts the inflected form of each lemma constraint based on the source context only. The inflected lemmas can then be incorporated into \nmt using any of the aforementioned constrained \nmt techniques.
Compared with \tla, this framework is more flexible, as it can be applied to diverse types of \nmt architectures and inflection modules, and facilitates fast adaptation to new terminologies without retraining the base \nmt model from scratch. This flexibility is enabled by the cross-lingual nature of the inflection module, which predicts the inflected form of each target lemma based on the source context only. This differs from traditional inflection models that predict the inflected forms based on pre-specified morphological tags or monolingual target context. 

Based on this framework, this paper makes the following contributions:
\begin{itemize}
    \item We construct and release test suites to evaluate models' ability to inflect terminology constraints for domain adaptation (English-German Health) and low-resource \mt (English-Lithuanian News).
    \item We show that integrating linguistic knowledge through a simple rule-based inflection module improves over its neural counterpart in intrinsic and end-to-end \mt evaluations.
    \item Our framework improves autoregressive and non-autoregressive translation, and outperforms the existing \tla approach for inflecting terminology translation. We open-source the code to facilitate replication and extensions. 
\end{itemize}

\section{Background}
\label{sec:background}

\paragraph{Autoregressive \nmt with Constraints}
Terminology constraints can be incorporated in autoregressive \nmt models via  
\begin{inparaenum}[1)]
    \item constrained decoding where constraint terms are incorporated in the beam search algorithm~\citep{HokampL2017,PostV2018}, or
    \item constrained training where \nmt models are trained to incorporate constraints using synthetic parallel data augmented with constraint terms on the source side~\citep{SongZYLWZ2019,DinuMFA2019}.
\end{inparaenum}
These approaches all assume that the constraints are provided in the correct inflected forms and can be directly copied to the target sentence. \citet{BergmanisP2021} extended the constrained training approach of~\citet{DinuMFA2019} to incorporate lemma-form constraints in an end-to-end way \---\ the inflected form of the lemma constraints are predicted jointly during translation. This approach requires a dedicated \nmt model architecture to integrate constraints as additional inputs to the encoder, and learns inflection solely from the parallel data. By contrast, our approach can be applied to multiple \nmt architectures and uses linguistically motivated rule that generalize better to rare and unseen terms.

\paragraph{Non-Autoregressive \nmt with Constraints}

Instead of generating the output sequence incrementally from left to right, non-autoregressive \nmt generates tokens in parallel~\citep{GuBXLS2018,Oord2018,MaZLNH2019} or by iteratively editing an initial sequence ~\citep{LeeMC2018,GhazvininejadLLZ2019}. Architectures differ with the nature of edit operations: the Levenshtein Transformer~\citep{GuWZ2019} relies on insertion and deletion, while \editor~\citep{XuC2021} uses insertion and reposition~(where each input token can be repositioned or deleted). Edit-based non-autoregressive generation provides a natural way to incorporate constraints in \nmt \---\ the constraints can be put into the initial sequence and edited to produce the final translation \citep{SusantoCT2020,XuC2021,WanKLCM2020}. 
Our approach can augment this family of techniques by inflecting constraints before they are used for further editing.

\paragraph{Morphological Inflection} 
Morphological inflection is the process of alternating the morphological form of a lexeme that adds morpho-syntactic information of the word in a sentence~(e.g. tense, case, number). Traditionally, morphological inflection as computational task is framed as predicting the inflected form of a word given its lemma and a set of morphological tags~(e.g. \textsc{n;acc;pl} represents a plural noun used in accusative case)~\citep{Cotterell2017SIGMORPHON}. The task was traditionally tackled using hand-engineered finite state transducer that relies on linguistic knowledge~\citep{Koskenniemi1983,Kaplan1994}, while recent work has shown impressive results by modeling it using neural sequence-to-sequence models~\citep{Faruqui2016}. More recently, a context-based inflection task has been proposed where the inflected form of a lemma is predicted given the rest of the sentence as context~\citep{Cotterell2018SIGMORPHON}. The state-of-the-art models for the task are neural models trained on supervised data~\citep{Cotterell2018SIGMORPHON,Kementchedjhieva2018}.
The inflection module in our framework differs from those for the context-based inflection task in that it requires cross-lingual context-based inflection \---\ it predicts the inflected form of a target lemma based only on the source language context. 

\paragraph{Morphologically-Aware Translation} 
In phrase-based \mt, modeling morphological compounds on the source~\citep{KoehnK2003} and target sides~\citep{Cap2014} improves translation quality. In \nmt, morphologically-aware segmentation is also useful when translating from or into morphologically complex languages~\citep{Huck2017,Ataman2018,Banerjee2018}. \citet{Tamchyna2017} propose to overcome data sparsity caused by inflection by training \nmt models to predict the lemma form and morphological tag of each target word. Different from prior work, we incorporate grammatical and morphological knowledge in an inflection module for terminology constraints in \nmt.
\section{Inflecting Target Lemmas Given the Source Context}
\label{sec:approach}

\begin{figure*}[!ht]
    \centering
    \begin{subfigure}[b]{0.32\textwidth}
        \centering
        \includegraphics[scale=0.36]{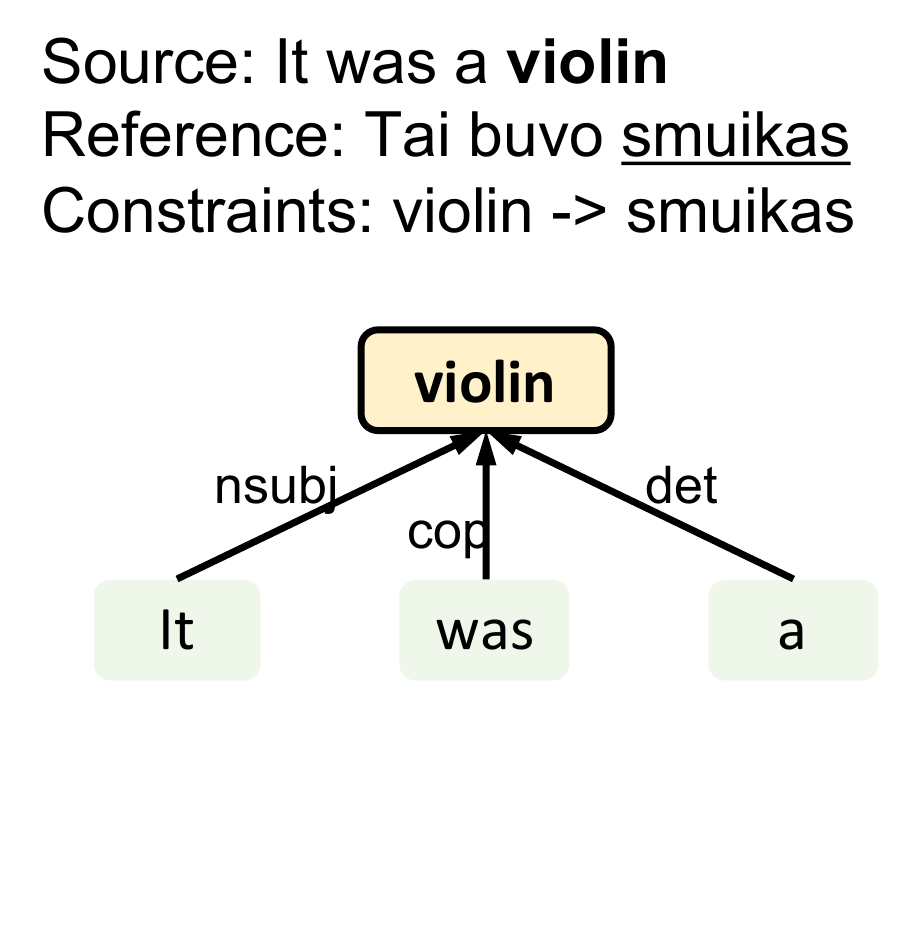}
        \caption{Nominative Case}
    \end{subfigure}
    \hspace{-5pt}
	\centering
    \begin{subfigure}[b]{0.32\textwidth}
        \centering
        \includegraphics[scale=0.36]{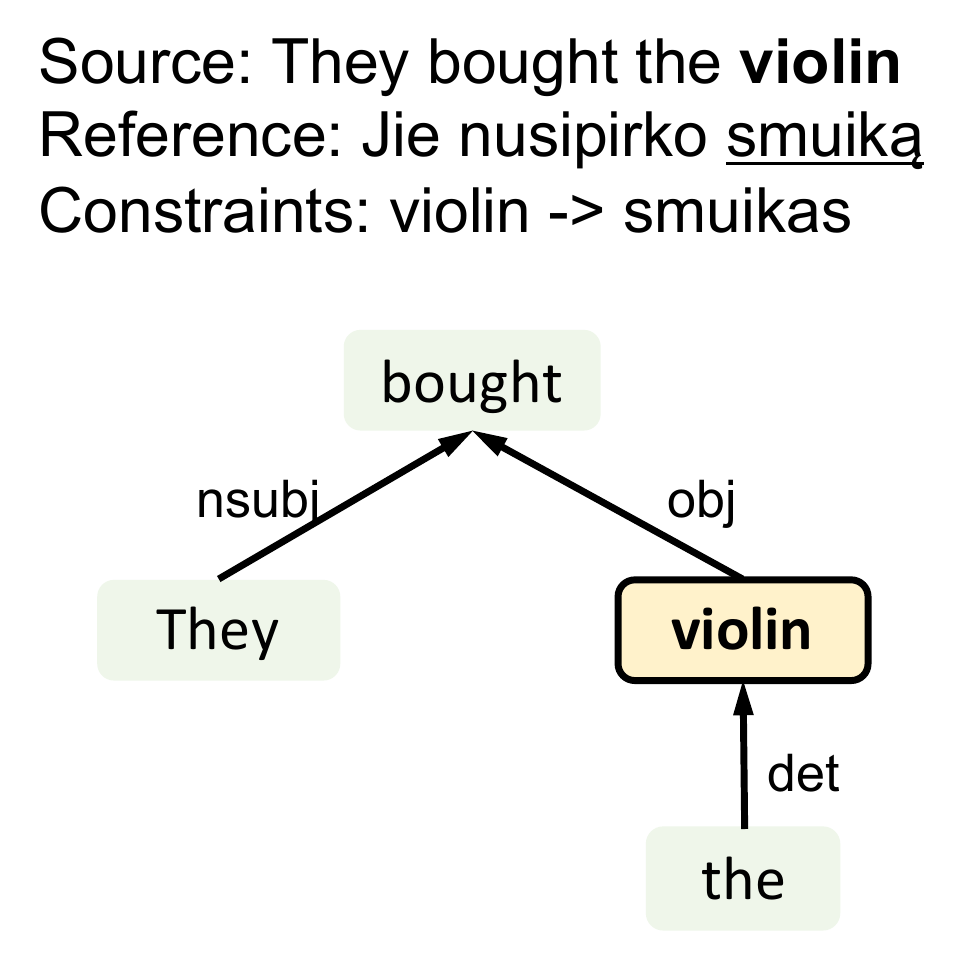}
        \caption{Accusative Case}
    \end{subfigure}
    \hspace{-5pt}
	\centering
    \begin{subfigure}[b]{0.32\textwidth}
        \centering
        \includegraphics[scale=0.36]{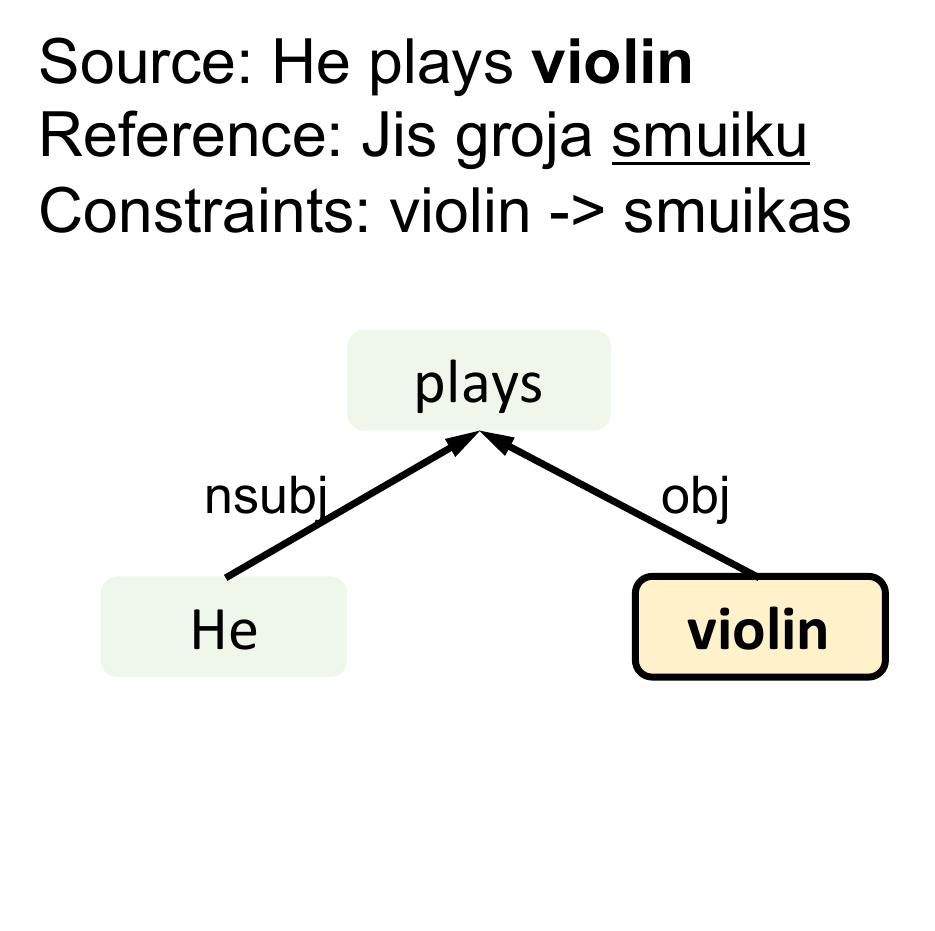}
        \caption{Instrumental Case}
    \end{subfigure}
\caption{Examples showing how the grammatical case of a target lemma is inferred from the dependency parsing tree of the source sentence. In each example, the reference usage of the target constraint is underlined, and its corresponding source term is boldfaced and highlighted in the yellow, outlined box in each dependency tree. Figure~(a) shows an example where the constraint term ``smuikas'' is used in \textit{nominative} case in the reference, since its the root in the dependency tree. In Figure~(b), the same constraint term is used in \textit{accusative} case in the reference, since it is the \textit{object} of the root verb ``bought''. However, not all objects should be used in \textit{accusative} case. As shown in Figure~(c), ``smuikas'' is used in \textit{instrumental} case, since it serves the \textit{instrument} with which the subject performs the action.}
\label{fig:deptree_example}
\vspace{-10pt}
\end{figure*}

\looseness=-1
We introduce a modular framework for inflecting terminology constraints for \nmt, where we first build an inflection module that predicts the inflected form of each target lemma term based on the source sentence and then incorporate the inflected constraints in \nmt using any of the aforementioned techniques. By framing the problem this way, we assume that the inflected forms can be inferred based only on the source context and integrated in a fluent translation by \nmt models. In cases where there are multiple possible inflected forms corresponding to different ways of translating the source, the inflection module can predict one of the possible forms, and the \nmt model can generate a translation conditioned on the predicted forms of the constraints.
Compared with \citet{BergmanisP2021}, our framework is more flexible \---\ it can be combined with any \nmt model that enables translation with constraints and can leverage diverse types of morphological inflection modules in which linguistic knowledge can be easily incorporated.

Formally, given a source sequence~$\boldsymbol{x}$ and~$k$ target lemma words~$\boldsymbol{\bar{z}} = (\bar{z}_1, \bar{z}_2, ..., \bar{z}_k)$ that need to be inflected, the inflection module~$\Theta$ predicts the inflected form of each target lemma~$\boldsymbol{z} = (z_1, z_2, ..., z_k)$ independently:
\begin{equation}
    p(\boldsymbol{z} \cond \boldsymbol{x}, \boldsymbol{\bar{z}}; \Theta) = \prod_{i=1}^{k} p(z_i \cond \boldsymbol{x}, \bar{z}_i; \Theta)
\end{equation}

\subsection{Rule-Based Inflection Module}
One can predict the inflected form of a target word given its lemma and the source context in two steps: first predict the morphological tag of the target word based on the source context, and then predict the inflected form based on the lemma and morphological tag. The second step can be modeled using traditional inflection models~\citep{Cotterell2017SIGMORPHON}, while the first step can be performed using rule-based inference based on linguistic knowledge. \citet{Mccarthy2020UniMorph} present a universal morphological~(UniMorph) paradigm with universal morphological tags for hundreds of world languages. In UniMorph, the morphological tag of a verb includes information about the tense~(past, present, or future), mood~(indicative, conditional, imperative, or subjunctive), the number~(singular or plural) and person~(first, second, or third person) of the subject. The tag of a noun or adjective includes information about gender~(masculine or feminine), number, and grammatical case. Some of these can be inferred from the target lemma~(e.g. the gender of a noun) or the source term~(e.g. the number of a noun), while some others need to be inferred based on the grammatical function of the source term in the sentence~(e.g. grammatical case) or the sentence-level semantics~(e.g. mood). Many of the inference rules are shared across a wide range of languages, except for the tense and mood of verbs, as well as the gender and some grammatical cases of nouns and adjectives.

In our rule-based inflection module, we extract the morphological features, part-of-speech tags, and dependency parsing tree of the source sentence using  pre-trained Stanza models\footnote{\url{https://github.com/stanfordnlp/stanza}} and infer the aforementioned classes based on grammar rules and validation examples. The tense and mood of a verb are inferred from the morphological form of the corresponding source term,\footnote{We ignore tense and mood types that cannot be inferred from the source term.} while the number and person of its subject are inferred based on the morphological form of its subject. For nouns and adjectives, the number can be inferred from the morphological form of the source term or modified noun, while the gender can be determined based on the target lemma.

To infer the grammatical case of a noun or adjective, one needs to infer about the grammatical role of the source term in the sentence. For example, in Lithuanian, there are seven main cases, including nominative, genitive, dative, accusative, instrumental, locative, and vocative cases. Figure~\ref{fig:deptree_example} shows examples of how the case of a Lithuanian noun can be inferred from the dependency parsing tree of the source sentence. Some of the cases can be easily distinguished from the others, while some are more difficult to infer. In this example, the nominative case is comparatively easy to infer \---\ the noun should be in the nominative case when the corresponding source term is the root or subject of the sentence. However, to distinguish between dative, accusative, instrumental, and locative cases, one needs to infer based on the grammatical and semantic role of the source term. In our rule-based module, we only take into account the most common scenarios.\footnote{Our code only includes a few simple inference rules written by non-expert based on the grammar knowledge from Wikipedia pages.}

Finally, given a lemma and its morphological tag, one can look up its inflected form in a morphological dictionary. We use DEMorphy~\citep{DEMorphy2018} for German and Wiktionary\footnote{\url{https://www.wiktionary.org/}} for Lithuanian. Since most Lithuanian nouns follow a set of declension rules,\footnote{\url{https://en.wikipedia.org/wiki/Lithuanian_declension}} we inflect Lithuanian nouns based on the rules for lemmas unseen in the dictionary.

\subsection{Neural Inflection Module}

As prior work shows that BERT-style architectures~\citep{Devlin2019BERT} can encode morphological information in their hidden representations and disambiguate morphologically ambiguous forms via contextualized encoding~\citep{Edmiston2020}, we build the neural-based inflection module as a substitution model and base it on the encoder-decoder Transformer architecture, which embeds the source sentence through the encoder and the target lemmas through the decoder. Next, the decoder predicts the inflected form of each target word in parallel. The inflection module resembles the architecture of the conditional masked language model~(\cmlm)~\citep{GhazvininejadLLZ2019} but differs in decoder input and output: \cmlm takes the target sentence with some tokens masked out as input and is trained to predict only the masked tokens conditioned on unmasked ones, while our inflection module takes target tokens in their lemma forms as input and predicts their inflected forms.

\cmlm only allows for one-to-one substitution of subwords. However, in the case of inflection, the number of subwords that constitute a lemma and its inflected form may differ. To facilitate varying-length substitution, we construct the decoder input by inserting~$K$ placeholders at the end of each target lemma. Next, the model predicts the token~$t \in \mathcal{V} \cup \{ \text{[PLH]} \}$ to be inserted at each input position. If~$t = \text{[PLH]}$, we delete the token at this position, otherwise we replace the token at this position with~$t$.\footnote{So for instance, given the input ``freeze [PLH] [PLH]'', the model could predict the output ``fro@@ zen [PLH]''.} 
\begin{table*}
\centering
\scalebox{0.85}{
\begin{tabular}{m{6em}m{14em}m{6em}m{14em}}
\toprule
 & Source & Constraints & Reference \\
\midrule
\multirow{1}{*}{En-De Health} & {The routine use of \textbf{abdominal} drainage to reduce postoperative complications after \textbf{appendectomy} for complicated \textbf{appendicitis} is controversial.} & {\textbf{abdominal} \underline{abdominell} \newline \textbf{appendectomy} \underline{Appendektomie} \newline \textbf{appendicitis} \underline{Appendizitis}} & {Die routinemäßige Verwendung von \underline{\textit{abdomineller}} Drainage zur Verminderung postoperativer Komplikationen nach einer \underline{Appendektomie} bei komplizierter \underline{Appendizitis} ist umstritten.} \\
\midrule
\multirow{2}[20]{*}{En-Lt News} & {A fire in 1939 left the building badly damaged, but as Father \textbf{Johnson}'s parishioners made plans to rebuild , they commissioned the \textbf{carillon}.} & {\textbf{Johnson} \underline{Johnsonas} \newline \textbf{carillon} \underline{karilionas}} & {1939 m. kilęs gaisras smarkiai apgadino pastatą, tačiau Tėvo \underline{\textit{Johnsono}} parapijiečiai planavo jį atstatyti, todėl užsakė \underline{\textit{karilioną}}.} \\\cmidrule{2-4}
 & {The expert who played the \textbf{carillon} in July called it something else: ``A cultural treasure'' and ``an irreplaceable historical instrument''.} & {\textbf{carillon} \underline{karilionas}} & {Liepos mėnesį \underline{\textit{karilionu}} grojęs ekspertas pavadino jį kitaip: ``kultūros lobiu'' ir ``nepakeičiamu istoriniu instrumentu''.} \\
\midrule
\end{tabular}
}
\caption{Examples from the English$\rightarrow$German~(En-De) health and English$\rightarrow$Lithuanian~(En-Lt) news test suites. For En-Lt, we select two examples from the same document. The annotated source terms are boldfaced and the target constraint terms are underlined. Some terms can be copied to the target~(e.g. ``Lymphödem'' and ``klinisch'' in En-De), while some others need to be inflected in the target sentence~(italicized).}
\label{tab:example}
\end{table*}

\begin{table}
\centering
\scalebox{0.95}{
\begin{tabular}{lrrr}
\toprule
 & \#Sent & \#Const & \#Const.Inf \\
\midrule
Health & 3000 & 4589 & 802 \\
News & 823 & 374 & 132 \\
\midrule
\end{tabular}
}
\caption{Number of sentences~(\textit{\#Sent}), constraints~(\textit{\#Const}), and constraints that need to be inflected~(\textit{\#Const.Inf}) in the health and news test suites.}
\label{tab:test_suite}
\vspace{-15pt}
\end{table}

\section{Evaluation Test Suites}
\label{sec:data}

To evaluate the models' ability to incorporate diverse types of lemma constraints in different context, we choose the two morphologically complex languages \---\ German and Lithuanian \---\ as the target languages, both of which are fusional languages with strong suffixing. We create two test suites \---\ the English$\rightarrow$German health test suite\footnote{To the best of our knowledge, there is no public health (or any non-news) domain \mt test set for English$\rightarrow$Lithuanian.} to evaluate models in the domain adaptation scenario and English$\rightarrow$Lithuanian news test suite to test models in the low-resource setting. Different from the automotive test suite of \citet{BergmanisP2021}, which contains short sentences~(15 tokens per source sentence on average) annotated with limited types of constraints~(mostly nouns and proper nouns), our test suites contains longer sentences~(20 and 25 tokens per source sentence on average) and diverse types of constraints including adjectives, nouns, proper nouns, and verbs. Different from the upcoming WMT21 terminology task\footnote{\url{http://statmt.org/wmt21/terminology-task.html}} where the terminology translation table includes different forms for a given source term, our test suites only provides terminology translations in lemma forms.

\paragraph{Health Test Suite}
We construct the health test suite to test the models' ability to integrate terminology translations for fast domain adaptation. The test set contains English health information text annotated with domain-specific terminology translations and the human-translated sentences in German.
We extract English$\rightarrow$German test examples from the Himl Test Set,\footnote{\url{http://www.himl.eu/test-sets}} which consists of English health information texts manually translated into German. We extract keyphrases from each source sentence using Yet Another Keyword Extractor~(\textsc{yake})~\citep{Campos2020YAKE}\footnote{\textsc{yake} extracts n-grams as keyphrases based on word casing, frequency, position, and their sentence context.} and filter out phrases with high or medium frequency in the training corpora since they are mostly common and domain-generic phrases.\footnote{We filter out keyphrases with frequency~$>100$ in the \wmt news training data.} We extract terminology translations from WikiTitles\footnote{\url{http://data.statmt.org/wikititles/v1/} and \url{http://data.statmt.org/wikititles/v2/}} and an online English-German dictionary,\footnote{\url{https://www.dict.cc/}} and annotate the keyphrases whose dictionary translations match the reference translation. 
As shown in Table~\ref{tab:example}, each source sentence in the test set is annotated with health-related terminology translations in the lemma forms, some of which can be directly copied to the final translation while some need to be inflected based on the context.

\paragraph{News Test Suite}
The news test suite simulates the scenario where a user looks up keyphrases of a document in a bilingual dictionary and pick the top translation for each keyphrase as a constraint to help low-resource \mt. We choose English$\rightarrow$Lithuanian as an example of low-resource translation. 
The test suite is constructed from English$\rightarrow$Lithuanian test examples from \wmt2019 news test sets. We first extract keyphrases from each source document using \textsc{yake}. Then, we find the top translation of each keyphrase (for many terms there's only one translation available) in an online dictionary.\footnote{\url{https://lithuanian.english-dictionary.help}} We filter out the keyphrases whose translations do not match the reference. Table~\ref{tab:example} shows two examples from the same document in the test suite. All occurrences of a keyphrase in one document are annotated with its target translation to encourage consistent translation of keyphrases within a document.\footnote{Interestingly, in Lithuanian, the masculine foreign names are usually translated by appending a suffix to the name to reflect their inflection forms. In this example, the foreign name ``Johnson'' is translated into ``Johnsonas'' in the nominative form in the dictionary, while in the reference it becomes ``Johnsono'' in the genitive form.}
Table~\ref{tab:test_suite} shows the number of sentences and constraints in each test suite.

\section{Experimental Settings}
\label{sec:exp_setup}

\paragraph{Training Data} 
\looseness=-1
For English$\rightarrow$German~(En-De), we use the training corpora from WMT14~\citep{BojarWMT2014} and \textit{newstest2013} for validation. For English$\rightarrow$Lithuanian, we use the training data from WMT19~\citep{BarraultWMT19} and \textit{newsdev2019} as the validation set. For preprocessing, we apply normalization, tokenization, true-casing, and BPE~\citep{SennrichHB16bpe}.\footnote{See preprocessing details and data statistics in Appendix.}

\paragraph{Baselines} 
We compare our model with the following baselines:
\begin{itemize}
    \item \textbf{Auto-Regressive (\ar) baseline} without integrating terminology constraints.
    \item \textbf{\ar with Constrained Decoding (\cd)} to incorporate hard constraints~\citep{Post18sacreBLEU}.
    \item \textbf{\ar with Target Lemma Annotation} (\tla) that integrates lemma constraints as an additional input stream on the source side~\citep{BergmanisP2021}.
    \item \textbf{Non-AutoRegressive (\nar) baseline} based on the \editor model~\citep{XuC2021}.
    \item \textbf{\nar with constraints (\narc)} that integrates constraints as the initial sequence in \editor without explicit inflection.
\end{itemize}

\paragraph{MT Models} 
All models are based on the \emph{base} Transformer~\citep{Vaswani2017}.\footnote{See more details in Appendix.} All models are trained with the Adam optimizer~\citep{KingmaB15} with initial learning rate of~$0.0005$ and effective batch sizes of~$32k$ tokens for \ar models and~$64k$ tokens for \nar models for maximum~$300k$ steps.\footnote{As shown in prior work~\citep{ZhouNG2019}, the batch sizes for training \nar models are typically larger than the \ar model.} We select the best checkpoint based on validation perplexity. \nar models are trained via sequence-level knowledge distillation~\citep{KimR2016}. 
For decoding, we use beam search with a beam size of~$4$ for \ar and \ar with \tla, while for \ar with \cd we use a beam size of~$20$ as suggested in prior work~\citep{PostV2018}. To enhance constraint usage in \nar models, we adopt the techniques by~\citet{SusantoCT2020}: we prohibit deletions on constraint tokens or insertions within the constraint segments.

\paragraph{Neural Inflection Model}  Its synthetic training data is derived from the \mt parallel data. We first lemmatise and part-of-speech tag the target sentences using Stanza. We then randomly select adjectives, verbs, nouns, and proper nouns from each target sentence and train the inflection module to predict their inflected forms based on their lemma forms and the source sentence. Following \citet{BergmanisP2021}, we draw the proportion of words selected in each target sentence randomly from the uniform distribution between~$(0, 0.4]$.
For training, we initialize its encoder parameters using the \nar baseline encoder and train it using Adam optimizer with a batch size of~$32k$ tokens for maximum~$200k$ steps.

\begin{table*}[ht]
\centering
\scalebox{0.9}{
\begin{tabular}{lcrrrrrr}
\toprule
 & \multirow{2}{*}{\bleu} & \multicolumn{3}{c}{Lemma Usage} & \multicolumn{3}{c}{Term Usage} \\
 & & All &	No Inf &	Inf & All &	No Inf &	Inf \\
\midrule
\textbf{En-De Health} \\
\ar baseline & 31.9 & 61.2 & 61.1 & 61.6 & 56.7 & 59.6 & 43.0 \\
\ar w/ \cd & 33.4 & \textbf{98.6} & \textbf{99.1} & 96.3 & 82.6 & \textbf{99.1} & 4.5 \\
\ar w/ \tla & \textbf{33.8} & 96.6 & 97.0 & 95.0 & \textbf{89.2} & 94.6 & 63.6 \\
\nar baseline & 31.0 & 56.1 & 56.4 & 54.7 & 52.8 & 55.2 & 41.3 \\
\narc & 31.1 & \textbf{99.0} & \textbf{99.1} & \textbf{98.5} & 82.0 & \textbf{99.1} & 1.4 \\
\midrule
\ar w/ \cd + neural & 33.3 & 95.6 & 95.9 & 91.1 & 81.0 & 91.1 & 33.3 \\
\ar w/ \tla + neural & 33.6 & 94.5 & 95.1 & 91.9 & 85.5 & 90.2 & 63.5 \\
\narc + neural & 30.9 & 95.6 & 95.8 & 94.9 & 81.1 & 91.1 & 33.8 \\
\midrule
\ar w/ \cd + rule & \textbf{33.7} & 96.8 & 96.8 & 97.0 & 87.3 & 95.0 & 51.0 \\
\ar w/ \tla + rule & \textbf{33.9} & 95.2 & 95.5 & 94.1 & 87.9 & 92.1 & \textbf{68.0} \\
\narc + rule & 31.7 & 97.1 & 97.0 & \textbf{97.5} & 87.1 & 95.0 & 49.5 \\
\midrule\midrule
\textbf{En-Lt News} \\
\ar baseline & 14.1 & 64.7 & 76.9 & 42.4 & 55.3 & 74.0 & 21.2 \\
\ar w/ \cd & 13.8 & 89.8 & \textbf{99.6} & 72.0 & 65.2 & \textbf{98.8} & 3.8 \\
\ar w/ \tla & \textbf{14.4} & 81.5 & 90.1 & 65.9 & 67.9 & 88.0 & 31.1 \\
\nar baseline & \textbf{14.3} & 59.4 & 69.0 & 41.7 & 52.7 & 67.8 & 25.0 \\
\narc & \textbf{14.3} & 89.8 & \textbf{99.2} & 72.7 & 64.7 & \textbf{98.3} & 3.0 \\
\midrule
\ar w/ \cd + neural & 13.5 & 82.4 & 85.1 & 77.3 & 57.2 & 75.2 & 24.2 \\
\ar w/ \tla + neural & 14.2 & 81.6 & 86.8 & 72.0 & 63.1 & 78.5 & 34.8 \\
\narc + neural & 14.0 & 83.7 & 88.0 & 75.8 & 58.0 & 77.7 & 22.0 \\
\midrule
\ar w/ \cd + rule & 13.9 & \textbf{93.0} & 97.5 & 84.8 & \textbf{75.9} & 94.2 & \textbf{42.4} \\
\ar w/ \tla + rule & \textbf{14.3} & 85.3 & 90.5 & 75.8 & 70.3 & 87.2 & 39.4 \\
\narc + rule & \textbf{14.3} & \textbf{93.3} & 97.1 & \textbf{86.4} & \textbf{75.7} & 94.2 & \textbf{41.7} \\
\midrule
\end{tabular}
}
\caption{\bleu, lemma, and term usage rates on the En-De health and En-Lt news test suites. For lemma and term usage, we report scores on all constraints~(\textit{All}), constraints that require no inflection~(\textit{No Inf}), and constraints that require inflection~(\textit{Inf}). We boldface the highest scores and their ties based on the paired bootstrap significance test~\citep{Clark2011} with~$p < 0.05$.}
\label{tab:results}
\vspace{-10pt}
\end{table*}

\paragraph{Evaluation}
We evaluate translation quality using sacreBLEU~\citep{Post18sacreBLEU}. To evaluate how well the translation preferences are incorporated in the translation outputs, we measure \textbf{lemma usage rate} by first lemmatising the translation output and then computing the percentage of lemma terms that appear in the lemmatised output. To evaluate whether the terms are inflected correctly, we measure \textbf{term usage accuracy} by matching each lemma constraint with its inflected form in the reference and computing the percentage of reference inflected terms that appear in the translation output. 

\section{Results and Discussion}

\paragraph{Intrinsic Inflection Accuracy}
\label{sec:model_based_result}

To evaluate the quality of the inflection modules, we first compare the inflection accuracy of neural-based and rule-based inflection modules against the term usage accuracy of the \tla model. The rule-based inflection module achieves higher inflection accuracy than the neural-based module on both test suites: the neural-based module obtains~$81.2\%$ accuracy on En-De health set and~$15.4\%$ accuracy on En-Lt news set, while the rule-based module achieves~$87.6\%$ accuracy on En-De and~$77.4\%$ accuracy on En-Lt. The rule-based module achieves close accuracy to \tla on En-De~($89.2\%$ term usage accuracy) and higher accuracy on En-Lt~($67.9\%$ term usage accuracy). 

To investigate why the neural-based inflection underperforms the rule-based one, we examine how the training and validation perplexity changes over the number of training epochs~(see Appendix). On both languages, the validation perplexity stops decreasing after a few training epochs~($10$ epochs for En-De and~$20$ epochs for En-Lt) while the training perplexity decreases very slowly. The final training perplexity remains at around~$5.1$ on En-De and~$5.7$ on En-Lt, which is high considering the number of possible inflection forms given a German or Lithuanian lemma. This indicates that the neural-based module does not learn generalizable inflection rules from the data effectively.

\paragraph{End-to-End \mt Evaluation}
\label{sec:rule_bsed_result}

Table~\ref{tab:results} shows the impact of rule-based and neural-based inflection modules on top of a range of \ar and \nar baselines. \nar baselines without constraints achieves competitive \bleu to the \ar baseline on En-Lt and slightly lower \bleu on En-De, as in ~\citet{XuC2021}. Given lemma constraints, \ar with \cd without inflection obtains lower term usage accuracy and lower \bleu than \ar with \tla, as in \citet{BergmanisP2021}. Similar to \ar with \cd, \narc without inflection obtains lower term usage and close or lower \bleu than \ar with \tla.

\looseness=-1
Adding rule-based inflection helps all models leverage lemma constraints more accurately.
On En-De, it significantly improves term usage accuracy of \ar with \cd by~$+4.7\%$ and \narc models by~$+5.1\%$.\footnote{All mentions of significance are based on the paired bootstrap test~\citep{Clark2011} with~$p < 0.05$.} On En-Lt, it significantly improves both the lemma usage rate and term usage accuracy of \ar with \cd~($+3.2\%$ on lemma usage and $+10.7\%$ on term usage) and \narc~($+3.5\%$ on lemma usage and $+11.0\%$ on term usage). Remarkably, it also improves the term accuracy of En-Lt \ar with \tla, which is already trained to inflect the target lemma constraints. When evaluating only on constraints that require inflection, the rule-based modules improves by ~$4.4$--$8.3\%$ on \tla,~$38.6$--$46.5\%$ on \cd, and~$38.7$--$48.1\%$ on \narc. As expected based on inflection accuracy results, rule-based modules outperform neural-based ones across the board. These improvements in term usage preserve or slightly improve \bleu.\footnote{The improvements on \bleu is statistically significant for \narc on En-De, but not for other models.}, as can be expected since the constraints only constitute a small portion of the tokens in the translation outputs. Overall, these results indicate that our proposed framework is model-agnostic and supports our hypothesis that the lemma constraints can be effectively inflected based on the source context alone.

We now compare our framework against \tla. Rule-based inflection combined with \narc achieves close lemma and term usage rates~($\Delta \le 2\%$) to \tla on En-De,~$+11.8\%$ higher lemma usage, and~$+7.8\%$ higher term usage accuracy on En-Lt~(the improvements are significant). On En-Lt, the largest improvements are on constraints that require inflection:~$+20.5\%$ on lemma usage and~$+10.6\%$ on term usage. Incorporating the constraints preserves translation quality, with no significant difference in \bleu. Overall, these results show the benefits of integrating linguistic knowledge via rule-based inflection over purely data-driven approaches. Our approach is also more adaptive, as \narc with rule-based inflection does not require re-training the whole \nmt model to incorporate new lemma terms. Instead, new terms can be incorporated by updating the morphological dictionary used in the inflection module.

\paragraph{Cost Trade-offs} Implementing the rule-based inflection module for the first target language (Lithuanian) took around~$6$ hours~(including the time for learning the grammar knowledge from Wikipedia) by a computer scientist without prior knowledge of the target language nor formal linguistics training. 
The second language (German) implementation took only~$3$ hours, since some rules are shared across languages. By contrast, the neural-based module was implemented in about~$3$ hours but took around~$38$ hours to train a single model for one language pair on 2 GeForce GTX 1080 Ti GPUs. While these numbers do not provide a controlled comparison, they highlight that the rule-based module is relatively simple to build, as it can be done for both languages in~$7$-$15$\% of the time required to train the neural model.

\begin{figure}[!t]
    \centering
    \begin{subfigure}[b]{0.33\textwidth}
        \centering
        \includegraphics[width=\textwidth]{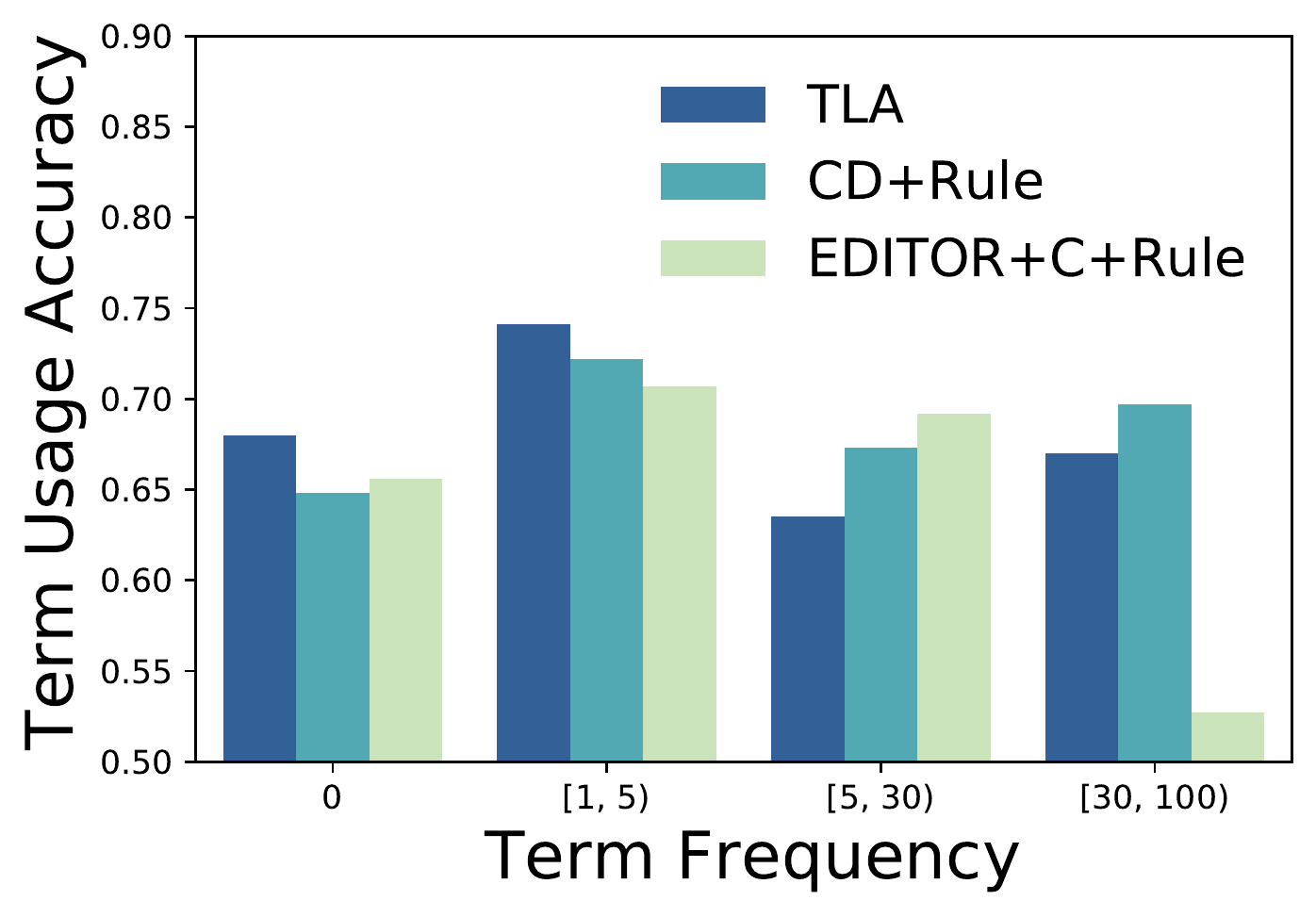}
        \caption{En-De}
    \end{subfigure}
	\centering
    \begin{subfigure}[b]{0.33\textwidth}
        \centering
        \includegraphics[width=\textwidth]{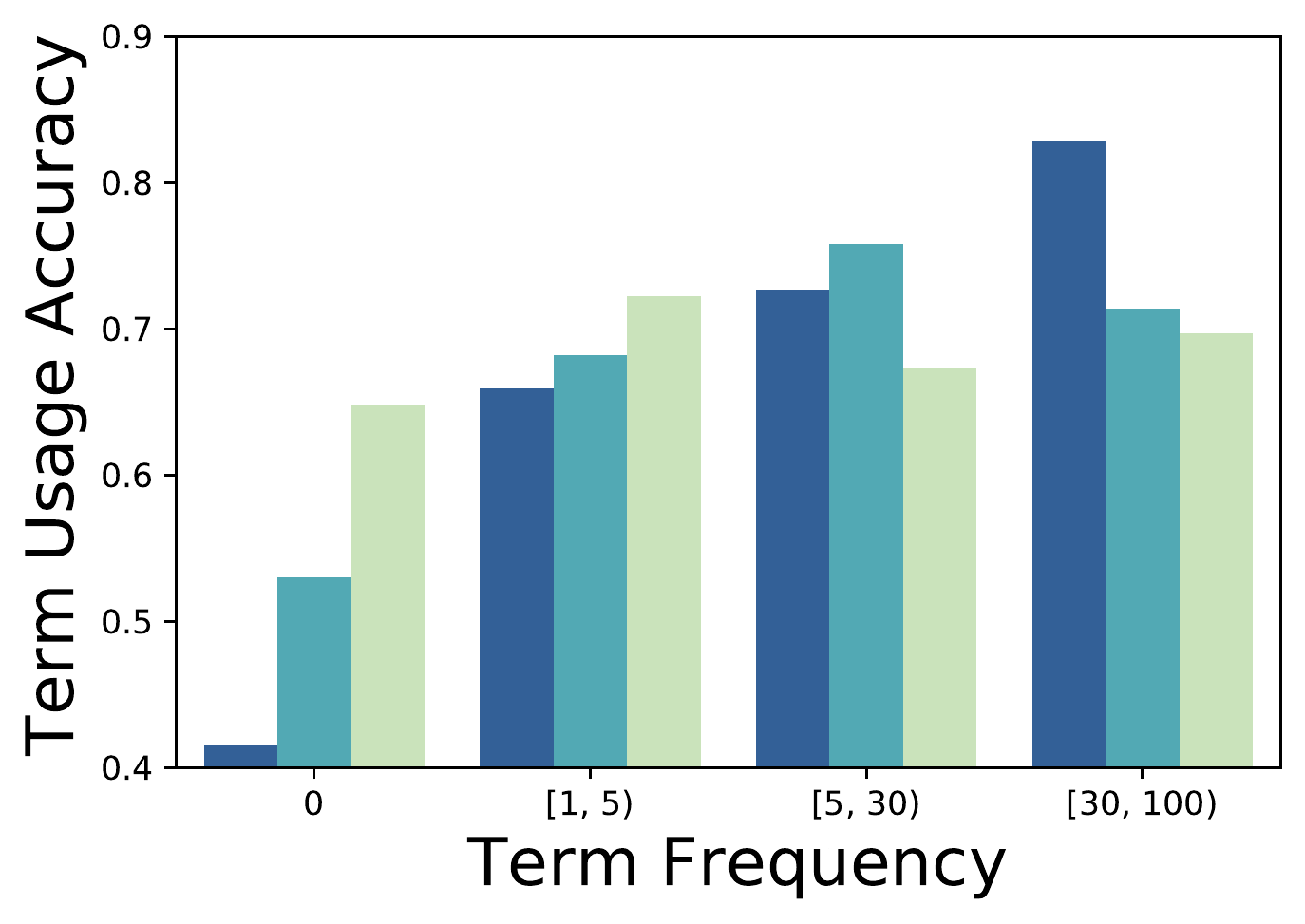}
        \caption{En-Lt}
    \end{subfigure}
	\hfill
\caption{Term usage accuracy of \tla, \cd + rule, and \narc + rule binned by training set frequency.}
\label{fig:term_usage_vs_freq}
\vspace{-10pt}
\end{figure}

\begin{table*}[t!]
\centering
\scalebox{0.95}{
\begin{tabular}{lm{28em}}
\toprule
source & {Jim Furyk's side need eight points from Sunday's 12 singles matches to retain the \textbf{trophy}.} \\
reference & {Jimo Furyko komandai reikia gauti aštuonis taškus sekmadienio 12 vienetų mačuose, kad išsaugotų \underline{trofėjų}.} \\
constraints & {\textbf{trophy}: \underline{trofėjus}} \\
reference inflection & {\textbf{trophy}: \underline{trofėjų}} (singular, accusative) \\
\tla & {Jim Furyk 's pusėje reikia aštuonių taškų iš sekmadienio 12 pažintys rungtynes išlaikyti \underline{trofėjus}.} \\
\tla + rule & {Jim Furyk 's pusėje reikia aštuonių taškų iš sekmadienio 12 pažintys rungtynes išlaikyti \underline{trofėjų}.} \\
\midrule
source & {In December 2017, he was accused of assaulting his father, Todd \textbf{Palin}.} \\
reference & {2017 m. gruodžio mėnesį jis buvo apkaltintas smurtu prieš savo tėvą Toddą \underline{Paliną}.} \\
constraints & {\textbf{Palin}: \underline{Palinas}} \\
reference inflection & {\textbf{Palin}: \underline{Paliną}} (singular, accusative) \\
\narc & {2017 m. gruodžio mėn. jis buvo apkaltintas užpuolimu jo tėvas Toddas \underline{Palinas}.} \\
\narc + rule & {2017 m. gruodžio mėn. jis buvo apkaltintas užpuolęs tėvą Toddą \underline{Paliną}.} \\
\midrule
\end{tabular}
}
\caption{Translation examples comparing \tla + rule against \tla, and \narc + rule against \narc on En-Lt. We boldface the source terms with translation constraints and underline the target constraint terms used in the reference and translation outputs.}
\label{tab:translation_example}
\vspace{-10pt}
\end{table*}

\paragraph{Term Frequency}

We analyze where rule-based inflection helps the most by computing the term usage accuracy on terms in different frequency bucket. As shown in Figure~\ref{fig:term_usage_vs_freq}, the trends are different on En-De and En-Lt. On En-De, \cd + rule slightly improves \tla on terms with frequency between~$[5, 100)$ instead of the rare terms. One reason is that the German morphological dictionary that we use to determine the gender of a word and its inflection forms only covers around~70\% of the constraint terms in the health test suite. In addition, \narc + rule underperforms \cd + rule on some constraint terms with frequency between~$[30, 100)$. This might be a side effect of knowledge distillation, which yields frequent errors for words that are rare in the training data~\citep{DingWLWTT2021}.
In En-Lt test set,~$68\%$ of the constraint terms are used in the inflection forms that are unseen in the training data. As shown in the figure, both \cd + rule and \narc + rule bring substantial improvements over \tla on terms that are unseen in the training data. This is because most Lithuanian nouns and adjectives are inflected based on a fixed set of rules, thus even when the target lemma is unseen in the training data or morphological dictionary, it can still be inflected correctly. As a result, the rule-based inflection module can effectively incorporate linguistic knowledge in translation models and thus generalizes better to rare and unseen terms.

\paragraph{Qualitative Analysis}
We examine a few randomly selected translation examples from \tla, \narc, and their counterparts with rule-based inflection.
As shown in Table~\ref{tab:translation_example}, \tla tends to copy constraint terms that are infrequent in the training data, and adding the rule-based inflection module helps \tla inflect the term correctly instead. In \narc models, the inflection module also improves the translation of the context around constraint terms, while the vanilla \narc model is prone to compounding errors caused by the uninflected constraints.

\section{Conclusion}
\looseness=-1
We introduced a modular framework for leveraging terminology constraints provided in lemma forms in neural machine translation. The framework is based on a novel cross-lingual inflection module that inflects the target lemma constraints given source context and an \nmt model that integrates the inflected constraints in the output. We showed that our framework can be flexibly applied to different types of inflection modules, including rule-based and neural-based ones, and different \nmt models, including autoregressive and non-autoregressive ones, with minimal training costs. Results on the English-German health and English-Lithuanian test suites showed that the linguistically motivated rule-based inflection module helps \nmt models incorporate terminology constraints more accurately than both neural-based inflection and the existing end-to-end approach to incorporating lemma constraints. This work opens future avenues for further improving the inflection module by combining linguistic knowledge with data-driven approaches. Future work is needed to explore the strengths and weaknesses of this framework for languages with a broader range of morphological properties.

\section*{Acknowledgement}
We thank Sweta Agrawal, Hal Daum\'e III, the anonymous reviewers, the CLIP lab at UMD for their helpful and constructive comments, and Yuxin Xiong for her help on error analysis of German outputs. This research is supported in part by the Amazon Web Services Machine Learning Research Award and by the Office of the Director of National Intelligence (ODNI), Intelligence Advanced Research Projects Activity (IARPA), via contract \#FA8650-17-C-9117. The views and conclusions contained herein are those of the authors and should not be interpreted as necessarily representing the official policies, either expressed or implied, of ODNI, IARPA, or the U.S. Government. The U.S. Government is authorized to reproduce and distribute reprints for governmental purposes notwithstanding any copyright annotation therein.

\bibliography{anthology,emnlp2021}
\bibliographystyle{acl_natbib}

\appendix

\section{Data Preprocessing}
For preprocessing, we apply normalization, tokenization, true-casing, and BPE~\citep{SennrichHB16bpe} with~$37,000$ and~$24,500$ merging operations for En-De and En-Lt. Table~\ref{tab:statistics} shows the provenance and statistics of the preprocessed data.

\section{Model and Training Details}
All models are based on the \emph{base} Transformer~\citep{Vaswani2017} with~$d_{\text{model}}=512$,~$d_{\text{hidden}}=2048$,~$n_{\text{heads}}=8$,~$n_{\text{layers}}=6$, and~$p_{\text{dropout}} = 0.3$. We tie the source and target embeddings with the output layer weights~\citep{PressW17,NguyenC18}.
We add dropout to embeddings~(0.1) and label smoothing~(0.1). All models are trained with the Adam optimizer~\citep{KingmaB15} with initial learning rate of~$0.0005$ and effective batch sizes of~$32k$ tokens for \ar models and~$64k$ tokens for \nar models for maximum~$300,000$ steps.\footnote{As shown in prior work, the batch sizes for training non-autoregressive models are typically larger than the \ar model~\citep{ZhouNG2019}.} We select the best checkpoint based on validation perplexity.
Following \citet{XuC2021}, we train \nar models using sequence-level knowledge distillation: we replace the reference sentences in the training data with translation outputs from the \ar models. To train the neural-based inflection module, we initialize its encoder parameters using the \nar baseline encoder and train it using Adam optimizer with a batch size of~$32k$ tokens for maximum~$200,000$ steps. Models are trained on 2 GeForce GTX 1080 Ti GPUs. Table~\ref{tab:model_size} shows the number of parameters in each model.
\begin{table}[h]
\centering
\scalebox{1}{
\begin{tabular}{lrrrr}
\toprule
 & Train & Valid & Provenance \\\hline
{En-De} & 3,961k & 3,000 & WMT14 \\
{En-Lt} & 1,612k & 1,964 & WMT19 \\
\midrule
\end{tabular}
}
\caption{Number of sentence pairs and provenance of the training and validation data.}
\label{tab:statistics}
\end{table}

\begin{table}[ht]
\centering
\begin{tabular}{lr}
\toprule
& Model Size (M) \\
\midrule
\textbf{En-De} \\
\ar & 65 \\
\ar w/ \cd & 65 \\
\ar w/ \tla & 65 \\
\nar  & 91 \\
rule-based inflection & 0 \\
neural-based inflection & 86 \\
\midrule
\textbf{En-Lt} \\
\ar & 57 \\
\ar w/ \cd & 57 \\
\ar w/ \tla & 58 \\
\nar & 84 \\
rule-based inflection & 0 \\
neural-based inflection & 72 \\
\hline
\end{tabular}
\caption{Model sizes (M) for the \ar, \nar, and inflection models.}
\label{tab:model_size}
\end{table}

\section{Evaluation Metric}
We evaluate translation quality using sacreBLEU~\citep{Post18sacreBLEU}.\footnote{BLEU+case.mixed+numrefs.1+smooth.exp+tok.13a\\+version.1.2.11}

\section{Learning Curves}
\begin{figure}[h!]
    \centering
    \begin{subfigure}[b]{0.42\textwidth}
        \centering
        \includegraphics[width=\textwidth]{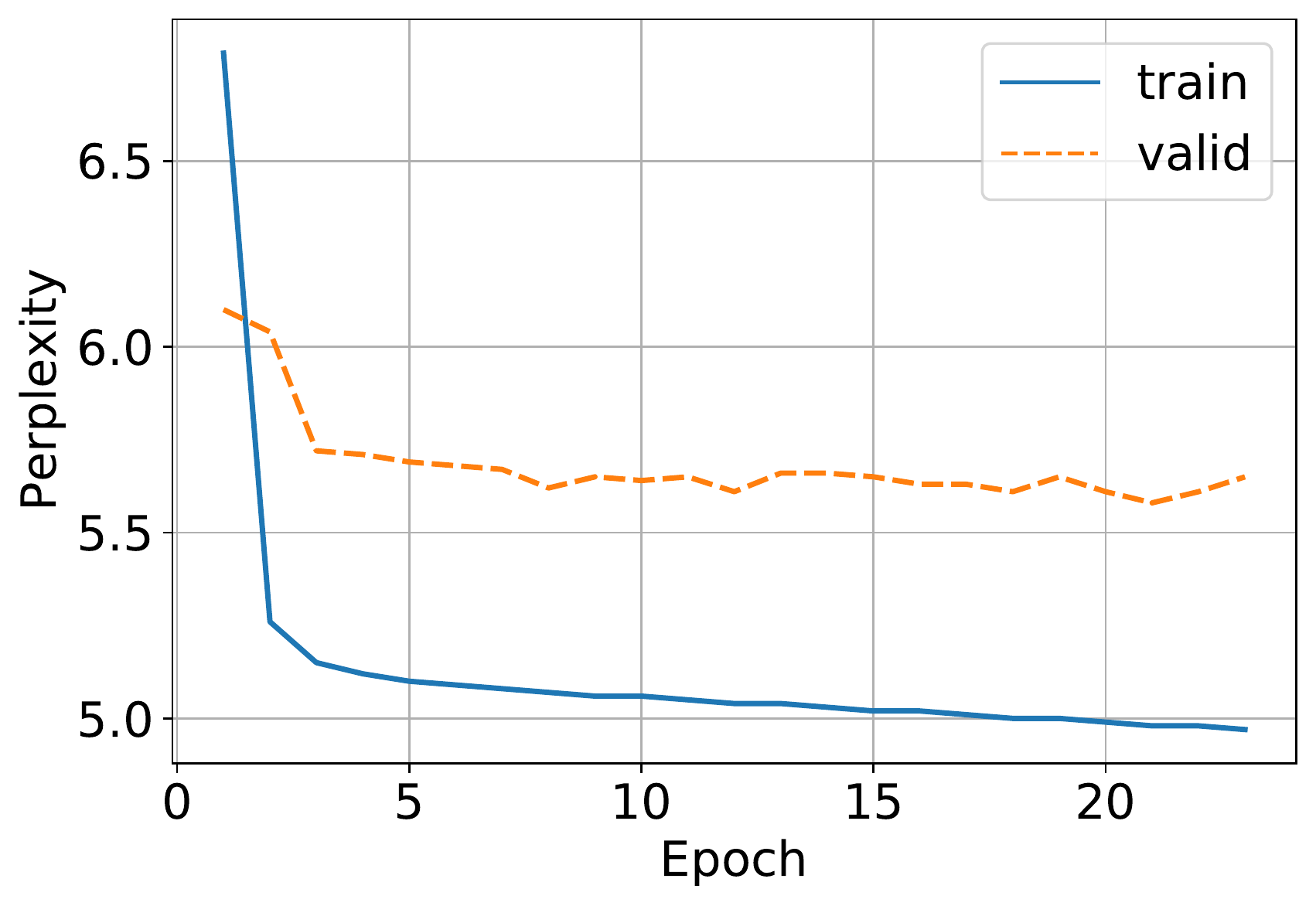}
        \caption{En-De}
        \label{fig:learning_curve_ende}
    \end{subfigure}
	\hfill
	
	\centering
    \begin{subfigure}[b]{0.42\textwidth}
        \centering
        \includegraphics[width=\textwidth]{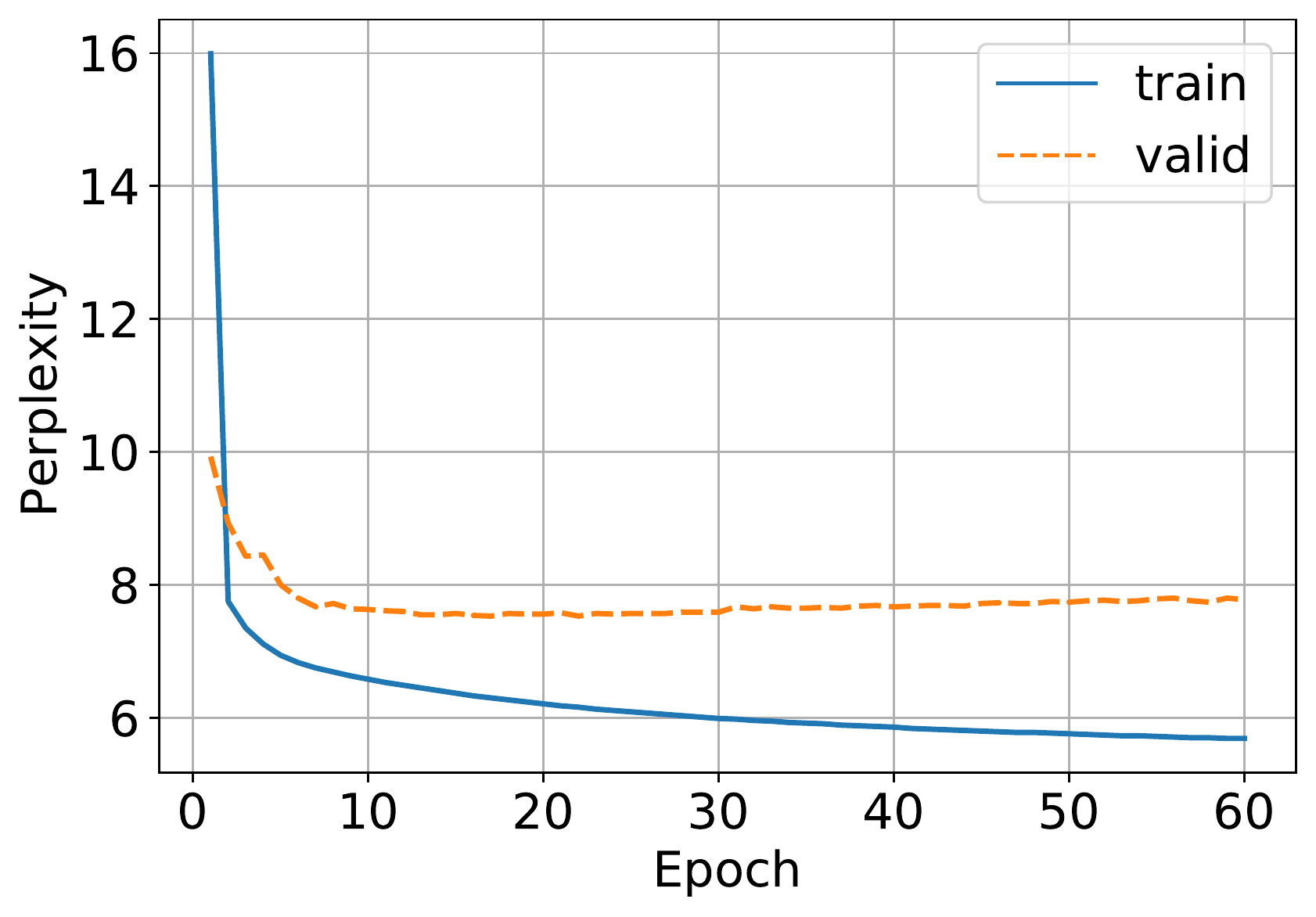}
        \caption{En-Lt}
        \label{fig:learning_curve_enlt}
    \end{subfigure}
	\hfill
\caption{Training and validation perplexity of the En-De and En-Lt neural-based inflection modules.}
\label{fig:learning_curve}
\end{figure}

Figure~\ref{fig:learning_curve} shows the learning curves of En-De and En-Lt neural-based inflection modules in terms training and validation perplexity.

\end{document}